\documentclass{article}

\PassOptionsToPackage{numbers, compress}{natbib}

\usepackage[final]{neurips_2025}

\usepackage[utf8]{inputenc} %
\usepackage[T1]{fontenc}    %
\usepackage{hyperref}       %
\usepackage{url}            %
\usepackage{booktabs}       %
\usepackage{amsfonts}       %
\usepackage{nicefrac}       %
\usepackage{microtype}      %
\usepackage{xcolor}         %
\usepackage{graphicx}       %
\usepackage{amsmath}
\usepackage{cleveref}
\usepackage{tabularx}       %
\usepackage{diagbox}
\usepackage{subcaption}
\usepackage{soul}
\usepackage{multirow}
\usepackage{enumitem}

\title{\emph{Smooth Reading}: Bridging the Gap of Recurrent LLM to Self-Attention LLM on Long-Context Tasks}

\author{%
Kai Liu$^{1,2}$ \and Zhan Su$^{3}$ \and Peijie Dong$^{4}$ \and Fengran Mo$^{3}$ \and Jianfei Gao$^{1}$ \and Shaoting Zhang$^{1}$ \and Kai Chen$^{1}$
\AND
$^{1}$Shanghai AI Laboratory \and $^{2}$Tongji University \and $^{3}$University of Montreal \and $^{4}$HKUST(GZ)\\
\AND
\texttt{\{liukai, gaojianfei, zhangshaoting, chenkai\}@pjlab.org.cn} \\
\texttt{\{zhan.su, fengran.mo\}@umontreal.ca} \\
\texttt{pdong212@connect.hkust-gz.edu.cn} \\
}

\usepackage{amsmath,amsfonts,bm}

\def\eqref#1{equation~\ref{#1}}

\def\1{\bm{1}}

\DeclareMathAlphabet{\mathsfit}{\encodingdefault}{\sfdefault}{m}{sl}
\SetMathAlphabet{\mathsfit}{bold}{\encodingdefault}{\sfdefault}{bx}{n}

\begin{document}

\maketitle

\begin{abstract}
    Recently, recurrent large language models (Recurrent LLMs) with linear computational complexity have re-emerged as efficient alternatives to self-attention-based LLMs (Self-Attention LLMs), which have quadratic complexity. However, Recurrent LLMs often underperform on long-context tasks due to their limited fixed-size memory. Previous research has primarily focused on enhancing the memory capacity of Recurrent LLMs through architectural innovations, but these approaches have not yet enabled Recurrent LLMs to match the performance of Self-Attention LLMs on long-context tasks. We argue that this limitation arises because processing the entire context at once is not well-suited for Recurrent LLMs. In this paper, we propose \emph{Smooth Reading}, a chunk-wise inference method inspired by human reading strategies. \emph{Smooth Reading} processes context in chunks and iteratively summarizes the contextual information, thereby reducing memory demands and making the approach more compatible with Recurrent LLMs. Our experimental results show that this method substantially narrows the performance gap between Recurrent and Self-Attention LLMs on long-context tasks, while preserving the efficiency advantages of Recurrent LLMs.
    Our \emph{Smooth Reading} boosts SWA-3B-4k (a Recurrent LLM) from 5.68\% lower to 3.61\% higher performance than Self-Attention LLMs on LongBench.
    Besides, our method maintains the high efficiency, training 3× faster and inferring 2× faster at 64k context compared to Self-Attention LLMs.
    To our knowledge, this is the first work to achieve comparable performance using Recurrent LLMs compared with Self-Attention LLMs on long-context tasks. We hope our method will inspire future research in this area. To facilitate further progress, we will release code and dataset.
\end{abstract}

\section{Introduction}
\label{sec:introduction}

Self-attention-based large language models (Self-Attention LLMs) have achieved remarkable success across a wide range of tasks, including those requiring long-context understanding~\citep{longbench, ruler}. The demand for improved long-context capabilities is increasing, driven by applications such as complex reasoning~\citep{deepseek-r1}, embodied agents~\citep{ebodied}, and deep research~\citep{deepresearcher}. However, the Self-Attention mechanism has quadratic computational complexity and linear space requirements, severely limiting its scalability to long input sequences.
In response, Recurrent LLMs have re-emerged as a promising alternative~\citep{rwkv4, mamba, deltaNet, gateddelta, TTT, GLA}. These models offer linear computational complexity and constant space usage, making them significantly more efficient than Self-Attention LLMs when processing long contexts.

Despite these efficiency advantages, current Recurrent LLMs still underperform compared to Self-Attention LLMs on long-context tasks~\citep{an_empirical_study_of_mamba}. A central limitation is their fixed memory capacity~\citep{GLA}, which restricts their ability to retain and utilize long-range information. This leads to a trade-off: while Recurrent LLMs are better suited for processing longer contexts due to their efficiency, their actual performance on long-context tasks remains inferior.
Recent research has focused on enhancing the memory capacity of Recurrent LLMs through architectural innovations~\citep{rwkv56, deltaNet, RWKV-7, mamba2, HGRN2, TTT}. For example,~\citep{rwkv56} increases memory size, while~\citep{deltaNet} introduces more expressive update rules to improve memory efficiency. Nevertheless, these architectural modifications have not yet enabled Recurrent LLMs to perform comparably with Self-Attention LLMs on long-context tasks. This situation motivates us to explore alternative strategies for improving Recurrent LLM performance in long-context scenarios.

\begin{figure}[t]
    \centering
    \begin{minipage}{0.445\textwidth}
        \centering
        \includegraphics[width=\textwidth]{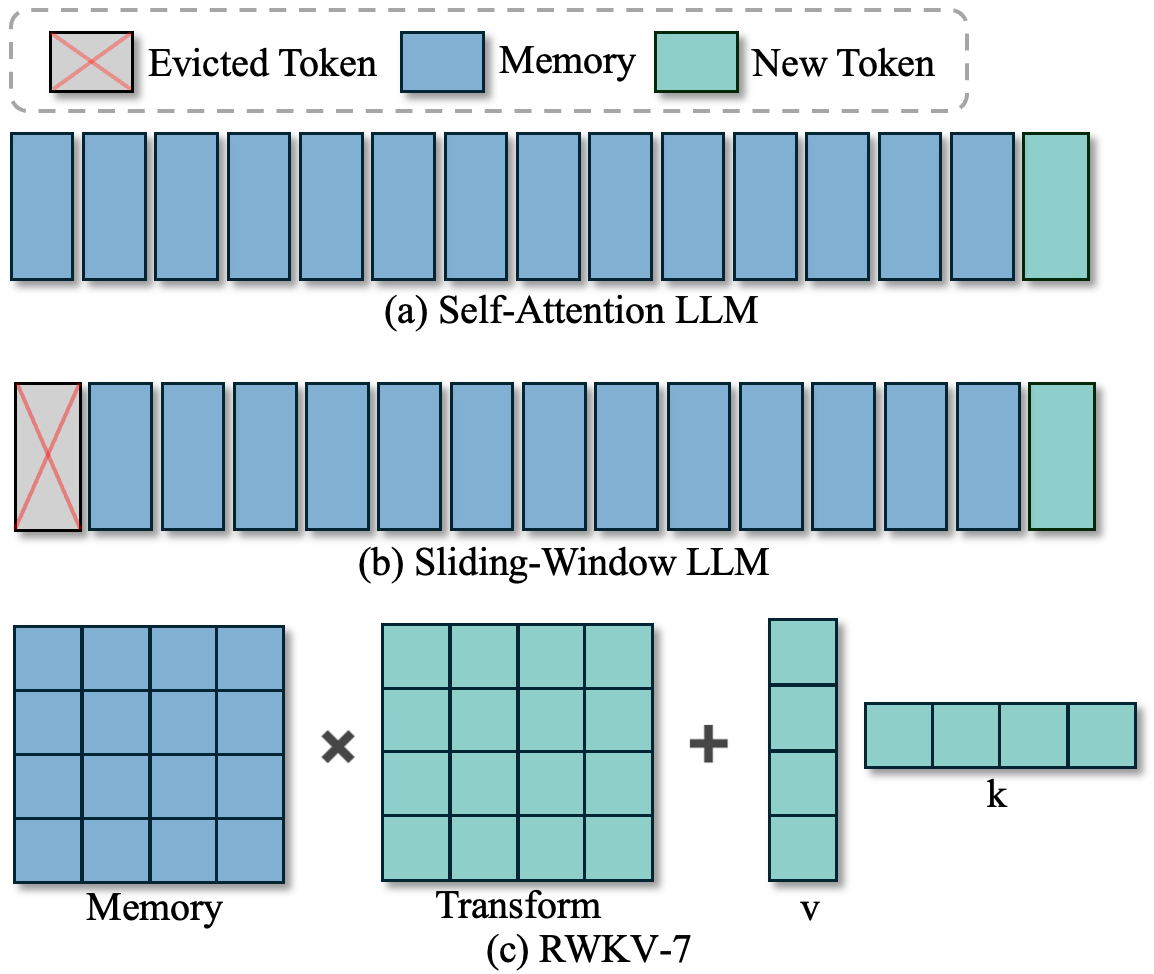}
        \caption{Comparison of Architectures: (a) Self-Attention LLM: A new token is appended to the memory. (b) Sliding-Window LLM: A new token is appended to the memory, and the earliest token is evicted. (c) RWKV-7: ``Transform'', ``k'', and ``v'' are generated from the new token and used to update memory. The memory size of Self-Attention LLM is virtually unlimited, while the other two are fixed-size.}
        \label{fig:arch}
    \end{minipage}%
    \hfill
    \begin{minipage}{0.485\textwidth}
        \centering
        \includegraphics[width=\textwidth]{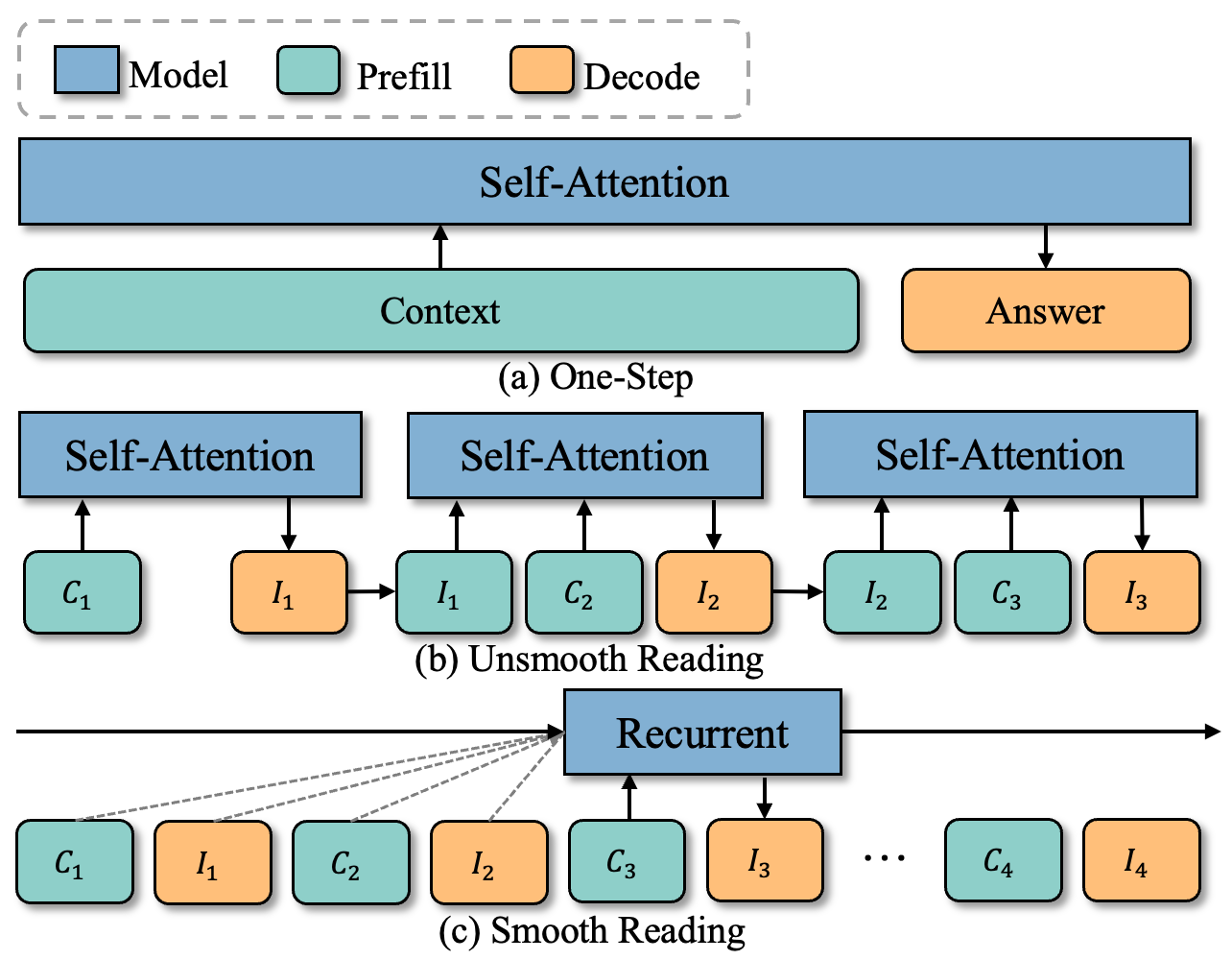}
        \caption{Comparison of Inference Methods for Long-context Tasks. (a) One-Step Inference: The entire context is processed at once. (b) Unsmooth Reading: The context is divided into chunks, and the model re-feeds the contextual summary at each step. (c) Smooth Reading: The model processes chunks iteratively, updating its hidden memory without re-feeding the contextual summary.}
        \vspace{0.3cm}
        \label{fig:infer}
    \end{minipage}
\end{figure}

We argue that \textit{a tailored inference method is essential for improving Recurrent LLMs on long-context tasks.} Previous works typically relied on One-Step inference, processing the entire context in a single pass (see \Cref{fig:infer}(a)). While effective for Self-Attention LLMs, which can attend to the whole sequence at once, this approach is less suitable for Recurrent LLMs, whose memory capacity is inherently limited. Inspired by the way humans read, sequentially scanning and integrating information, we introduce Smooth Reading, a novel inference strategy designed specifically for Recurrent LLMs. As illustrated in \Cref{fig:infer}(c), \emph{Smooth Reading} divides the input context into smaller chunks and processes them incrementally. At each step, the model reads a chunk, generates a contextual summary, and stores salient information in its hidden memory, thereby avoiding overwhelming the model with the entire long context at once. This strategy allows Recurrent LLMs to process long contexts smoothly and efficiently, providing an inference path better matched to their capabilities. Specifically, we curate a new dataset to train Recurrent LLMs specifically for the \emph{Smooth Reading} procedure. Unlike architectural improvements, our method enhances Recurrent LLMs through an optimized inference process that offers explicit guidance on information importance. Notably, our technique is orthogonal to architectural modifications and can be used in conjunction with them.

We evaluate our method using transformer-based LLMs with sliding-window attention (Sliding-Window LLMs)~\citep{longformer} and RWKV-7~\citep{RWKV-7} as representative Recurrent LLMs, and test on two long-context benchmarks: Needle-in-a-Haystack (NIAH)~\citep{ruler} and LongBench~\citep{longbench}.
Our results show that Recurrent LLMs trained with \emph{Smooth Reading} achieve performance comparable to, or even exceeding, that of Self-Attention LLMs on long-context tasks. Specifically, our SWA-3B-4k-SR outperforms Self-Attention LLMs by an average of 3.61\% on LongBench. Moreover, Recurrent LLMs with \emph{Smooth Reading} retain core advantages of recurrent architectures:
(1) Length extrapolation: Recurrent LLMs are expected to generalize to contexts much longer than those seen during training. Our method can inherit the length extrapolation ability of Recurrent LLMs.
For example, our SWA-3B-4k-SR, trained on sequences up to 32k tokens, can extrapolate its performance to at least 256k tokens.
(2) Efficiency: Our method preserves the linear computational complexity of Recurrent LLMs, resulting in significant speed-ups. On a 64k context, our SWA-3B-4k-SR is approximately 3$\times$ faster in training and 2$\times$ faster in inference compared to the corresponding Self-Attention LLM.

Our main contributions are as follows:
\begin{itemize}[leftmargin=15pt]
    \item We propose \emph{Smooth Reading}, a novel inference method for Recurrent LLMs, and curate a new dataset specifically designed to train models for this purpose.
    \item We demonstrate that \emph{Smooth Reading} significantly improves the long-context performance of Recurrent LLMs, achieving results on par with Self-Attention LLMs. To the best of our knowledge, this is the first work to close the gap using Recurrent LLMs.
    \item We show that our approach preserves the inherent advantages of Recurrent LLMs, including superior length extrapolation and computational efficiency.
\end{itemize}

\section{Related Work}

\subsection{Architecture of LLMs}

\textbf{Self-Attention LLMs}: Self-attention mechanisms~\citep{attention} form the foundation of most widely used LLMs, such as those described in~\citep{deepseek-r1,qwen2.5,gpt4}. In this architecture, all previous tokens are retained in memory. When a new token is processed, it is appended to the existing memory, as illustrated in \Cref{fig:arch}(a). This design enables Self-Attention LLMs to maintain virtually unlimited memory capacity; however, it also incurs quadratic computational complexity with respect to sequence length.

\textbf{Recurrent LLMs}: Recurrent LLMs maintain a fixed-size memory, resulting in linear computational complexity and constant space usage. Most contemporary Recurrent LLMs use variants of linear attention~\citep{linear_attention}, such as RWKV-7~\citep{RWKV-7}, where new token information is integrated into memory via addition, as shown in \Cref{fig:arch}(c). Additionally, LLMs employing sliding-window attention (Sliding-Window LLMs)~\citep{longformer} can also be considered recurrent. These models achieve linear complexity and constant space usage by discarding the earliest tokens once the stored sequence exceeds the window size, as depicted in \Cref{fig:arch}(b).
The fixed memory size of Recurrent LLMs inherently limits their memory capacity, leading to reduced performance on long-context tasks~\citep{an_empirical_study_of_mamba}. To address this limitation, various architectural advancements have been proposed, which can be classified into two groups: (1) increasing memory size~\citep{HGRN2,mamba,mamba2,mom}, and (2) improving memory efficiency~\citep{deltaNet,TTT,gateddelta}. Despite these improvements, Recurrent LLMs still underperform compared to Self-Attention LLMs on long-context tasks.

We improve Recurrent LLMs by optimizing the inference method—without modifying their architecture. This approach offers two key benefits: (1) it guides the model on how to prioritize and update information, and (2) it dynamically adjusts the number of tokens used for aggregation, enabling longer context summaries when necessary. Since our method works independently of architectural changes, it remains compatible with existing innovations and can be combined for further gains.

\begin{figure}[t]
    \centering
    \includegraphics[width=.85\textwidth]{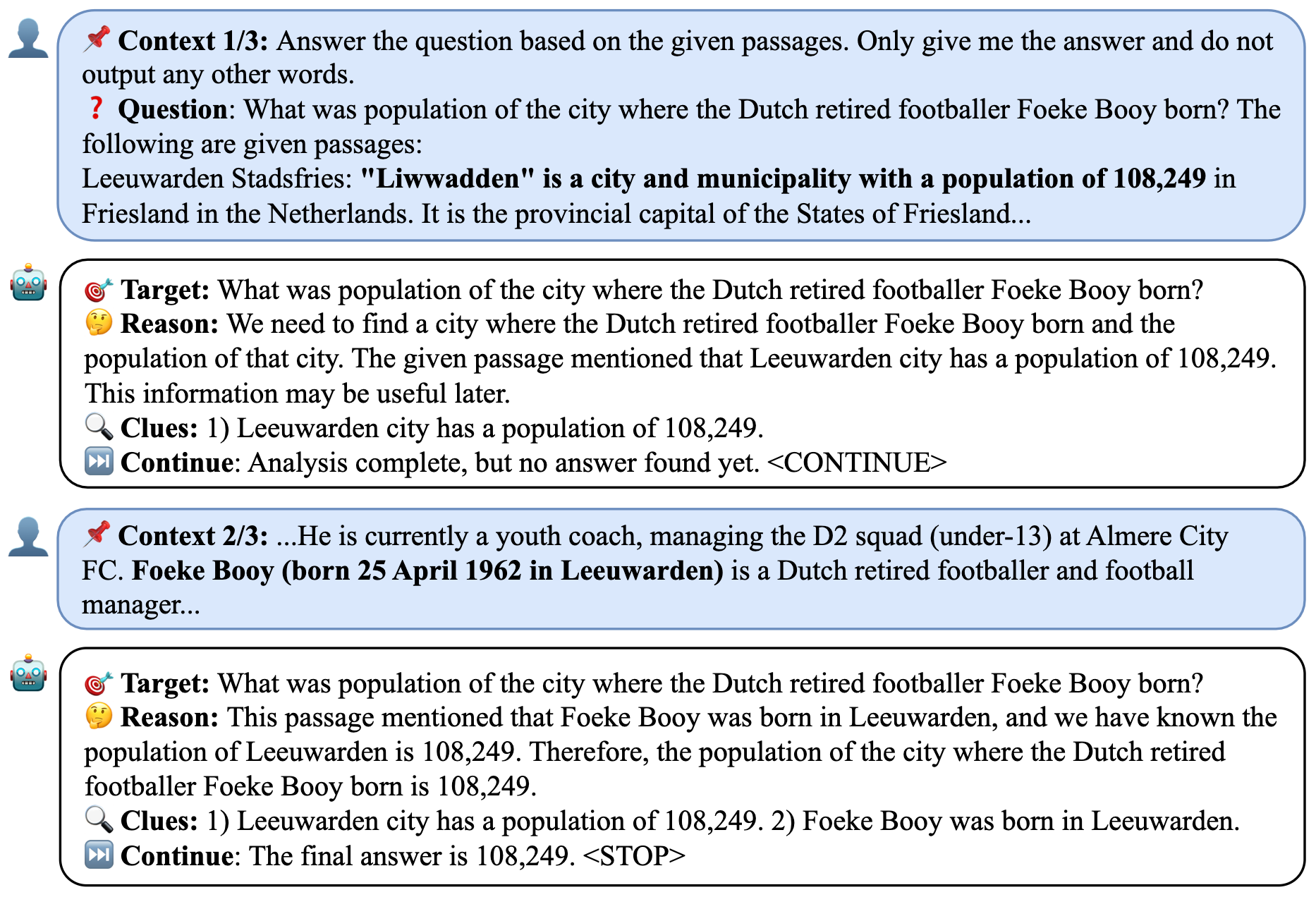}
    \caption{Illustration of \emph{Smooth Reading}. White boxes indicate the information gathered by the model and the decision points for continuing reading. Blue boxes represent chunks of the query context.}
    \label{fig:active_reading}
    \vspace{-0.3cm}
\end{figure}

\subsection{Inference Methods for Long-Context Tasks}

\textbf{One-Step Inference}: This is the most commonly used approach, in which the entire context is processed in a single forward pass to generate an answer, as shown in \Cref{fig:infer}(a). This method requires the model to handle long contexts and maintain a large memory capacity. Thus, One-Step inference is suitable for Self-Attention LLMs but presents significant challenges for Recurrent LLMs.

\textbf{Multi-Step Inference}: To address the context length limitations of Self-Attention LLMs, Multi-Step inference methods have been proposed. For instance, Retrieval-Augmented Generation (RAG)~\citep{mog} divides the context into chunks and retrieves the most relevant ones to answer a query. Similarly, recent approaches~\citep{chain_of_agents,CompAct,are_long_llms_a_necessity} iteratively process long contexts by extracting relevant information from each chunk to form the final answer.
However, these methods typically require resetting the hidden memory at each step and re-feeding the accumulated information into the model to avoid the quadratic computational complexity that arises with increasing context length. This process can result in information loss. Therefore, we term these methods \emph{Unsmooth Reading}, as shown in \Cref{fig:infer}(b).

Our approach is specifically tailored for Recurrent LLMs, which can retain hidden memory across steps without incurring quadratic computational costs. This design enables a \emph{Smooth Reading} process, leading to more effective and efficient inference.

\section{Methodology}
\label{sec:method}

\begin{table}[t]
    \centering
    \caption{Comparison of Architecture and Inference Method Combinations. "Satisfying MR" indicates whether the memory requirements of inference methods are met by the corresponding architectures. "Hidden Memory" shows whether hidden memory can be maintained during inference. Disadvantages in each combination are \underline{underlined}.}
    \resizebox{.9\linewidth}{!}{
        \begin{tabular}{ll|cc|cc}
            \toprule        Architecture & Inference        & Computation          & Space  & Satisfying MR  & Hidden Memory  \\
            \midrule
            Self-Attention               & One-Step         & \underline{$O(L^2)$} & $O(L)$ & Yes            & --             \\
            Recurrent                    & One-Step         & $O(L)$               & $O(1)$ & \underline{No} & --             \\
            \midrule
            Self-Attention               & Unsmooth Reading & $O(L)$               & $O(1)$ & Yes            & \underline{No} \\
            Recurrent                    & Smooth Reading   & $O(L)$               & $O(1)$ & Yes            & Yes            \\
            \bottomrule
        \end{tabular}}
    \label{tab:related}
    \vspace{-0.1cm}
\end{table}

\subsection{Smooth Reading}
\label{sec:smooth_reading}

As discussed previously, different inference methods impose varying demands on a model's memory capacity. For recurrent LLMs with limited memory, it is crucial to develop tailored inference methods to enhance their performance on long-context tasks. We first analyze two inference methods—One-Step inference and Unsmooth Reading—designed for Self-Attention LLMs and then introduce our proposed method, Smooth Reading, specifically for recurrent LLMs.

We define the memory capacity demands of an inference method as \emph{memory requirement (MR)}, which can be estimated by the length of the sequence processed in a single step. Let \( f \) denote an LLM, which can be either a Self-Attention LLM \(f_S\) or a Recurrent LLM \(f_R\). The model receives a sequence and hidden memory, denoted by \(M\), as inputs and outputs a new sequence along with updated hidden memory.

\textbf{One-Step Inference}:
The standard approach processes the entire context at once with empty hidden memory:
\begin{align}
    A, M = f_S(C, \emptyset), \quad \mathrm{MR}_{f_S(C, \emptyset)} \propto \mathrm{len}(C||A) \nonumber
\end{align}
where \(C\) is the context, \(A\) is the answer, \(\mathrm{len}(\cdot)\) denotes the sequence length, and \(||\) indicates sequence concatenation. In One-Step inference, the complete context is processed at once with empty hidden memory (\(\emptyset\)). As the context length increases, the memory requirement grows correspondingly. While this is feasible for Self-Attention LLMs with linearly scaling memory, it is unsuitable for Recurrent LLMs with fixed memory capacity.

\textbf{Unsmooth Reading}:
To overcome the context window limitation and quadratic complexity of Self-Attention LLMs, a multi-step inference method, which we term Unsmooth Reading \citep{are_long_llms_a_necessity,chain_of_agents,CompAct}, was proposed.
This approach divides the context into chunks of maximum size \(c\): \(C = [C_1, C_2, \ldots, C_n]\), where \(\mathrm{len}(C_i) \leq c\) for each chunk \(C_i\).
To accumulate information across the context, Unsmooth Reading introduces a contextual summary \(I\), which aggregates information over the chunks. At each step \(i\), \(I_i\) is updated from the previous summary and the current chunk, as shown in \Cref{fig:infer}(b). Unsmooth Reading can be formulated as:
\begin{align}
    I_{i}, M = f_S(I_{i-1}||C_i, \emptyset), \quad \mathrm{MR}_{f_S(I_{i-1}||C_i, \emptyset)} \propto \mathrm{len}(I_{i-1} || C_i || I_i) \leq 3c
    \label{eq:system}
\end{align}
assuming \(\mathrm{len}(I_{i-1}), \mathrm{len}(C_i), \mathrm{len}(I_i) \leq c\). This bounds the per-step memory usage.
While Unsmooth Reading is designed for Self-Attention LLMs, it does not retain hidden memory across steps in order to avoid increasing memory size and quadratic computational complexity with respect to context length. The hidden memory \(M\) is reset at each step, resulting in information loss. Instead, the model must re-input the contextual summary \(I\) at each step, leading to an unsmooth reading process.
Therefore, although Unsmooth Reading significantly reduces the memory requirement, it is still suboptimal for Recurrent LLMs.

\textbf{Smooth Reading}: With recurrent LLMs, we can implement a \emph{Smooth Reading} inference method. Since recurrent LLMs have linear computational complexity and constant space usage, their inference efficiency is independent of context length. Thus, we can maintain hidden memory without incurring additional computation or memory costs. As the information in the contextual summary \(I\) is already stored in the model's hidden memory during its generation, we can directly use the hidden memory to replace the contextual summary \(I\) for the next iteration. Our process is formulated as:
\begin{align}
    I_{i}, M_{i} = f_R(C_i, M_{i-1})
    \label{eq:ours}
\end{align}
where \(M_{i-1}\) is the hidden memory from the previous step, and \(M_i\) is the updated hidden memory after processing chunk \(C_i\). Compared with Unsmooth Reading, \emph{Smooth Reading} maintains hidden memory across steps, allowing the model to retain more information. Furthermore, \emph{Smooth Reading} avoids repeatedly reading the contextual summary \(I\), resulting in better efficiency. As illustrated in \Cref{fig:infer}(c), \emph{Smooth Reading} interleaves the reading of chunks with the generation of contextual summaries, without additional intermediate steps.

A comparison of different architectures and inference strategies is presented in \Cref{tab:related}. Our results show that recurrent LLMs with \emph{Smooth Reading} offer the most efficient and effective combination, while other pairings are either less efficient or less effective. We provide our chunking method in \S \ref{sec:chunk} for reference.

\subsubsection{Content of Contextual Summary}

As presented in Figure~\ref{fig:active_reading}, the contextual summary \(I\) is a key component of our approach, ensuring that the model consistently accesses relevant local information. The summary comprises the following four elements:

\begin{itemize}[leftmargin=15pt]
    \item \textbf{Target}: To keep the model focused on the task objective and prevent distraction by irrelevant details, we explicitly record the task target.
    \item \textbf{Clues}: We maintain a collection of clues relevant to the task. For instance, in summarization tasks, this involves keeping an updated summary of all previously read chunks; in question answering, it consists of information directly related to the query.
    \item \textbf{Reason}: At each step, we document the rationale for any updates made to the clues.
    \item \textbf{Continue}: The model determines whether to proceed to the next chunk based on the accumulated information. If there is not enough information to answer the query, the model signals to continue reading by outputting a special token ``<CONTINUE>''. Once sufficient information has been gathered, the model outputs the final answer and signals completion with a ``<STOP>'' token.
\end{itemize}

In essence, the contextual summary must include all the key elements necessary for the model to answer the query effectively.

\subsubsection{Dataset Construction}
\label{sec:dataset_contruction}

To enable our models to learn \emph{Smooth Reading}, we construct a supervised fine-tuning (SFT) dataset based on two benchmarks: Needle-in-a-Haystack (NIAH)~\citep{ruler} and LongBench~\citep{longbench}. The tasks in these benchmarks are divided into several categories, and we create a dedicated dataset for each category. The dataset construction process for each task involves the following steps:

\begin{itemize}[leftmargin=15pt]
    \item \textbf{Raw Dataset}: For each task, we start by collecting the raw data from the training split of existing datasets, which includes query, answer, and context.
    \item \textbf{Teacher Model}: We employ a teacher model to simulate the \emph{Smooth Reading} process on the raw data. Depending on the complexity of the task, we use either a state-of-the-art (SOTA) LLM (deepseekv3~\citep{deepseekv3}) or rule-based models. For the LLM teacher, we implement Unsmooth Reading (\Cref{eq:system}) to collect data, as the LLM is self-attention-based and has quadratic complexity.
    \item \textbf{Early Stopping}: For certain tasks (e.g., question answering), we allow the teacher model to determine whether to continue reading or stop. If the model chooses to stop, we terminate the reading process and output the final answer; otherwise, the model processes all chunks.
    \item \textbf{Data Cleaning}: We clean the generated dataset using the same metric that is used for evaluating each task.
\end{itemize}

In total, we collect 48,856 data items. More details about dataset construction are provided in \S\ref{sec:dataset_construction_appendix}. An illustrative example of our \emph{Smooth Reading} process is shown in \Cref{fig:active_reading}. We use this dataset to train our models to predict the contextual summary \(I\) based on the provided chunks, thereby internalizing the ability of \emph{Smooth Reading} in the model.

\subsection{Inference Efficiency Analysis}
\label{sec:inference}

Our method introduces additional computational overhead—specifically, more generated tokens—to Recurrent LLMs. This raises an important question: does a Recurrent LLM with \emph{Smooth Reading} remain more efficient than a Self-Attention LLM utilizing One-Step inference?

Let \(p_r\) and \(d_r\) denote the wall-clock time per token for prefilling and decoding, respectively.
In this case, both \(p_r\) and \(d_r\) are constant and independent of the context length. Hence, we can use \(\beta \times p_r\) to replace \(d_r\), where \(\beta = \frac{d_r}{p_r}\).
Define \(c\) as the chunk size, \(n\) as the number of chunks, and \(l = c \times n\) as the total context length.
Let \(g\) denote the decoding length at each step. We assume \(g\) is a constant and independent of the context length, which is reasonable since the information needed usually relates to the query and is not affected by the context length.
The overall inference time for our method can be expressed as:
\begin{align}
    T_{\text{Recurrent-SR}} & = n \times c \times p_r + n \times g \times d_r = \left(1 + \frac{g \times \beta}{c}\right) \times l \times p_r \nonumber
\end{align}
Therefore, Recurrent LLMs with our method remain linear in complexity with respect to context length. They are still more efficient than Self-Attention LLMs with quadratic complexity on long contexts. Moreover, the formula suggests that increasing the chunk size and employing more efficient Recurrent LLMs can further enhance the efficiency of our approach.

\section{Experiments}

\begin{table}[t]
    \centering

    \caption{Results on LongBench. ``Infer'' indicates inference method. ``OS'' denotes One-Step inference, ``UR'' denotes Unsmooth Reading, and ``SR'' denotes Smooth Reading. The best results are highlighted in bold.}
    \resizebox{\textwidth}{!}{
        \begin{tabular}{l|l|c|cccccc|c}
            \toprule
            Architecture                    & Infer               & Model          & SQA            & MQA            & Summary        & FewShot        & Synthetic      & Code           & Avg            \\
            \midrule
            \multirow{2}{*}{Self-Attention} & OS                  & Qwen-2.5-3B-OS & 24.20          & 41.25          & \textbf{30.22} & 65.89          & 56.75          & 66.00          & 47.38          \\
            \cmidrule{2-10}
                                            & UR                  & Qwen-2.5-3B-UR & \textbf{31.54} & 42.17          & 21.91          & 68.96          & 61.50          & 64.13          & 48.37          \\
            \midrule
            \multirow{4}{*}{Recurrent}      & \multirow{2}{*}{OS} & RWKV-7-3B-OS   & 16.96          & 11.39          & 29.16          & 67.72          & 60.50          & 62.84          & 41.43          \\
                                            &                     & SWA-3B-4k-OS   & 16.43          & 26.14          & 26.96          & 66.43          & 48.00          & \textbf{66.22} & 41.70          \\
            \cmidrule{2-10}
                                            & \multirow{2}{*}{SR} & RWKV-7-3B-SR   & 28.87          & 40.02          & 28.23          & 65.90          & 65.25          & 59.92          & 48.03          \\
                                            &                     & SWA-3B-4k-SR   & 30.46          & \textbf{47.67} & 26.27          & \textbf{69.60} & \textbf{66.75} & 65.18          & \textbf{50.99} \\
            \bottomrule
        \end{tabular}
    }
    \label{tab:longbench}
    \vspace{-0.2cm}
\end{table}

\subsection{Experimental Details}
\label{sec:implementation_in_main}

We compare our Recurrent LLMs with \emph{Smooth Reading} against three alternative strategies: (1) a Self-Attention LLM with One-Step inference, (2) a Recurrent LLM with One-Step inference, and (3) a Self-Attention LLM with Unsmooth Reading.

\noindent\textbf{Models.} For Self-Attention LLMs, we use Qwen2.5~\citep{qwen2.5} as a representative model. For Recurrent LLMs, we consider two types:
(1) Sliding-Window LLMs: These are constructed by applying the sliding-window attention mechanism to the Qwen2.5 model. We refer to these as "SWA-\(x\)k," where \(x\) denotes the window size. Unless specified otherwise, we use a window size of 4k tokens.
(2) RWKV-7~\citep{RWKV-7}: a SOTA Recurrent LLM using a variant of linear attention.

\noindent\textbf{Training Dataset.} All models are trained on our curated dataset. In addition to the \emph{Smooth Reading} format, we prepare two alternative forms: (1) a One-Step form that includes only the context and answers, and (2) an Unsmooth Reading form. Since all three dataset variants share identical contexts and answers, the comparison remains strictly fair.
More details about the training and evaluation setups are provided in \S\ref{sec:training} and \S\ref{sec:eval}, respectively.

\subsection{Comparison of Long-Context Performance}
\label{sec:performance}

To assess the long-context performance of our method, we evaluate it on the LongBench~\citep{longbench} and NIAH~\citep{ruler} benchmarks, as shown in \Cref{tab:longbench} and \Cref{tab:ruler}. For NIAH, we focus on contexts ranging from 8k to 32k tokens, which aligns with our training length of 32k.

\textbf{One-Step inference: Recurrent LLMs underperform compared to Self-Attention LLMs.}
On LongBench, Qwen-2.5-3B-OS achieves an average accuracy of 47.38\%. In contrast, RWKV-7-3B-OS and SWA-3B-4k-OS both achieve less than 42\% on average.
On NIAH, Qwen-2.5-3B-OS reaches 98.13\% accuracy on average, while RWKV-7-3B-OS and SWA-3B-4k-OS achieve below 95\% on average.
These results indicate that One-Step inference is not suitable for Recurrent LLMs due to their limited memory capacity, which cannot hold the entire context at once.

\textbf{Smooth Reading significantly enhances the performance of Recurrent LLMs, making them competitive with Self-Attention LLMs.}
On LongBench, RWKV-7-3B-SR and SWA-3B-4k-SR achieve average accuracies of 48.03\% and 50.99\%, surpassing Qwen-2.5-3B-OS by 0.65\% and 3.61\%, respectively.
Similarly, on NIAH, RWKV-3B-SR and SWA-3B-4k-SR achieve performance comparable to Qwen-2.5-3B-OS.
These results demonstrate that \emph{Smooth Reading} is better suited to Recurrent LLMs and can substantially improve their long-context task performance.

\textbf{Recurrent LLMs with Smooth Reading outperform Self-Attention LLMs with Unsmooth Reading due to hidden memory.}
On LongBench, when comparing Qwen-2.5-3B-UR and SWA-3B-4k-SR—both derived from the same pretrained model (Qwen-2.5-3B)—SWA-3B-4k-SR yields an average accuracy that is 2.62\% higher.
On NIAH, Qwen-2.5-3B-UR shows relatively poor and unstable performance, which we attribute to the lack of hidden memory; it also lags behind SWA-3B-4k-SR and RWKV-3B-SR.
These findings suggest that the hidden memory in Recurrent LLMs is more effective than repeatedly inputting a contextual summary.

To further validate our method, we conducted experiments with models of 7B parameters. Refer to \S\ref{sec:7b} for details.
Additionally, we compare our method with more inference methods, as detailed in \S\ref{sec:rag}.
Overall, these results provide strong evidence of the effectiveness of our approach.

\subsection{Length Extrapolation}
\label{sec:length_extrapolation}

\begin{table}[t]
    \centering
    \caption{Results on the Needle-in-a-Haystack (NIAH). The context lengths range from 8k to 256k, while the training length is 32k.}
    \resizebox{\textwidth}{!}{
        \begin{tabular}{l|l|c|ccc|c|ccc|c}
            \toprule
            Architecture                    & Infer               & Model          & 8k             & 16k             & 32k             & Avg            & 64k             & 128k           & 256k           & Avg            \\
            \midrule
            \multirow{2}{*}{Self-Attention} & OS                  & Qwen-2.5-3B-OS & 98.80          & 98.60           & 97.00           & 98.13          & 9.00            & 0.00           & 0.00           & 3.00           \\
            \cmidrule{2-11}
                                            & UR                  & Qwen-2.5-3B-UR & 87.80          & 87.00           & 92.60           & 89.13          & 93.80           & 95.00          & 67.20          & 85.33          \\
            \midrule
            \multirow{4}{*}{Recurrent}      & \multirow{2}{*}{OS} & RWKV-3B-OS     & 98.40          & 95.80           & 86.60           & 93.60          & 39.00           & 8.60           & 0.00           & 15.87          \\
                                            &                     & SWA-3B-4k-OS   & 53.60          & 22.40           & 11.60           & 29.20          & 6.80            & 1.80           & 1.60           & 3.40           \\
            \cmidrule{2-11}
                                            & \multirow{2}{*}{SR} & RWKV-3B-SR     & 99.40          & 98.80           & 97.20           & 98.47          & 75.20           & 20.60          & 1.94           & 32.58          \\
                                            &                     & SWA-3B-4k-SR   & \textbf{99.80} & \textbf{100.00} & \textbf{100.00} & \textbf{99.93} & \textbf{100.00} & \textbf{99.80} & \textbf{99.60} & \textbf{99.80} \\
            \bottomrule
        \end{tabular}
    }\label{tab:ruler}
    \vspace{-0.3cm}
\end{table}

We further evaluate the length extrapolation capability of our method on NIAH~\citep{ruler}. As shown in \Cref{tab:ruler}, all models are trained with a 32k context length. As detailed in \S\ref{sec:rwkv-7}, it has been proven that Sliding-Window LLMs possess strong length extrapolation ability, whereas RWKV-7 exhibits weak extrapolation.

\textbf{Self-Attention LLMs}: One-Step inference achieves high accuracy up to the training length (32k) but fails to generalize to longer contexts. For example, the accuracy of Qwen-2.5-3B-OS drops to 0\% at 64k. Unsmooth Reading improves the ability of Self-Attention LLMs to handle longer contexts, but performance remains unsatisfactory: Qwen-2.5-3B-UR never exceeds 95\% accuracy at any length.

\textbf{Sliding-Window LLMs}: With One-Step inference, Sliding-Window LLMs demonstrate gradual performance degradation as context length increases, indicating some degree of extrapolation capability, albeit at a low performance level—for instance, SWA-3B-4k-OS reaches only 53.6\% at 8k. In contrast, when equipped with \emph{Smooth Reading}, Sliding-Window LLMs improve substantially: SWA-3B-4k-SR achieves 99.6\% accuracy at 256k, demonstrating strong length extrapolation ability.

\textbf{RWKV}: RWKV's length extrapolation ability is limited. Its performance drops sharply beyond the training length when using One-Step inference, falling from 86.6\% at 32k to 39.0\% at 64k. With our \emph{Smooth Reading} method, RWKV-3B-SR performs better at extended lengths (75.2\% at 64k), but still shows limited extrapolation, as accuracy drops to 1.94\% at 256k.

Overall, these results indicate that our method can inherit the length extrapolation capabilities of Recurrent LLMs, making it potentially scalable to extremely long contexts.

\subsection{Efficiency Comparison}
\label{sec:efficiency}

\begin{figure}
    \centering
    \begin{subfigure}{0.48\textwidth}
        \includegraphics[width=.95\textwidth]{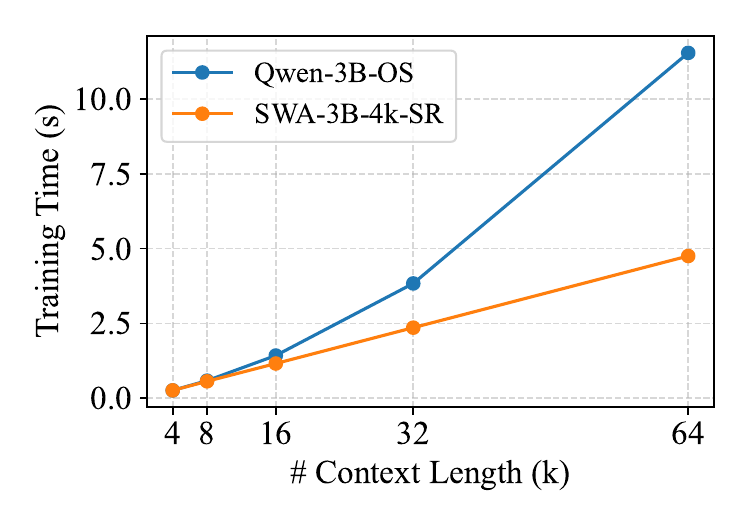}
        \subcaption{Training Efficiency}
    \end{subfigure}
    \begin{subfigure}{0.48\textwidth}
        \includegraphics[width=.95\textwidth]{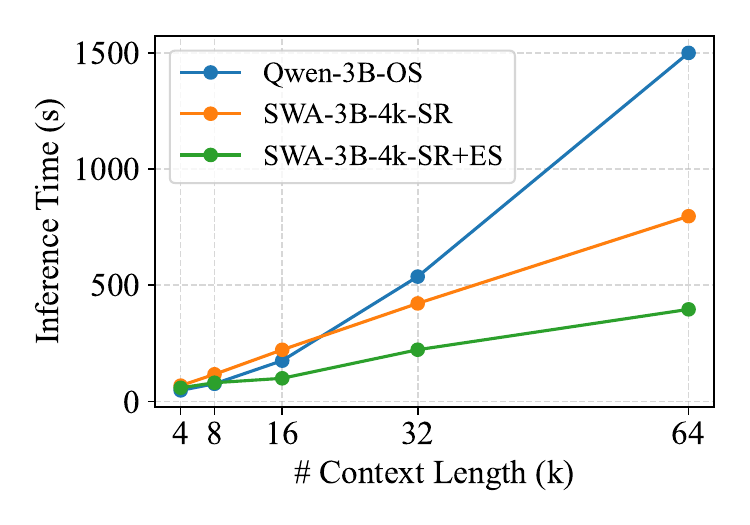}
        \subcaption{Inference Efficiency}
    \end{subfigure}
    \caption{Efficiency Comparison across Context Lengths (x-axis) vs. Inference Time (y-axis). Recurrent LLMs with \emph{Smooth Reading} scale linearly, while Self-Attention LLMs show quadratic growth. '+ES' indicates that early stopping is enabled in our method. The total token count scales with context length.}
    \label{fig:efficiency}
    \vspace{-0.4cm}
\end{figure}

\begin{table}[t]

    \centering
    \caption{Ablation Study on Window Size (W) and Chunk Size (C) for Accuracy (\%) and Inference Time (s) on NIAH.}
    \begin{subtable}{0.46\textwidth}
        \centering
        \caption{Performance Comparison}
        \begin{tabular}{l|cccc}
            \toprule
            \diagbox{W}{C} & 512   & 1024  & 2048  & 4096 \\
            \midrule
            512            & 97.0  & 77.8  & 0.0   & 0.0  \\
            1024           & 99.2  & 89.0  & 26.4  & 12.6 \\
            2048           & 100.0 & 100.0 & 95.4  & 31.2 \\
            4096           & 99.8  & 99.8  & 100.0 & 83.4 \\
            \bottomrule
        \end{tabular}
    \end{subtable}
    \hfill
    \begin{subtable}{0.46\textwidth}
        \centering
        \caption{Inference Time Comparison}
        \begin{tabular}{l|cccc}
            \toprule
            \diagbox{W}{C} & 512 & 1024 & 2048 & 4096 \\
            \midrule
            512            & 528 & 457  & 327  & 100  \\
            1024           & 537 & 444  & 374  & 364  \\
            2048           & 577 & 471  & 387  & 343  \\
            4096           & 646 & 505  & 423  & 366  \\
            \bottomrule
        \end{tabular}
    \end{subtable}
    \label{tab:window_chunk}
    \vspace{-0.2cm}
\end{table}

A principal advantage of Recurrent LLMs is their linear computational complexity with respect to context length, in contrast to the quadratic complexity of Self-Attention LLMs. In this section, we compare the efficiency of both training and inference for these models. For Recurrent LLMs, we use the Sliding-Window LLM as a representative model, since an optimized inference engine for RWKV-7 is currently unavailable.

\textbf{Training Efficiency}: Sliding-Window LLMs consistently demonstrate superior training efficiency. As shown in \Cref{fig:efficiency}(a), the training time for Sliding-Window LLMs remains significantly lower than that for Self-Attention LLMs. For instance, at a context length of 64k, the training time required by SWA-3B-4k-SR is approximately one-third that of Qwen-3B-OS.

\textbf{Inference Efficiency}: We evaluate inference time on the Needle-in-a-Haystack task across various context lengths. While Sliding-Window LLMs with \emph{Smooth Reading} process tokens faster, they incur some overhead from generating extra tokens, which makes them slightly less efficient than Self-Attention LLMs for short contexts. However, as context length grows, Sliding-Window LLMs with \emph{Smooth Reading} become significantly more efficient. As shown in \Cref{fig:efficiency}(b), our SWA-3B-4k-SR model halves the inference time compared to Qwen-2.5-3B-OS at a 64k context length.

Moreover, our \emph{Smooth Reading} enables flexible handling of long contexts by allowing early stopping once sufficient information is obtained. When early stopping is enabled, SWA-3B-4k-SR uses only a quarter of the inference time required by Qwen-2.5-3B-OS at a 64k context length.
These results highlight that our method is significantly more efficient than Self-Attention LLMs for long-context scenarios, making it a better fit for tasks requiring long-context processing.
Besides, our results show that early stopping has minimal impact on performance, as detailed in \S\ref{sec:early_stop}.

\subsection{Ablation Study}
Taking Sliding-Window LLM as an example, we conduct an ablation study on how memory size (simulated by window size) and chunk size impact performance and efficiency in Recurrent LLMs. Results for various window-chunk combinations are shown in \Cref{tab:window_chunk}.

\textbf{Chunk Size}: Increasing the chunk size can overload the model with too much information, leading to drops in performance. As shown in Table~\ref{tab:window_chunk}, with a window size of 512 and a chunk size of 4096, the accuracy falls to just 0.0\%. However, larger chunk sizes reduce the total number of processing rounds and generated tokens, which improves efficiency. For instance, at a window size of 4096, increasing the chunk size from 512 to 4096 reduces inference time from 646 to 366 seconds.

\textbf{Window Size}: Increasing the window size expands the model’s memory, allowing it to accommodate more information and process larger chunks. For example, with a chunk size of 2048, accuracy jumps from 0.0\% with a window size of 512 to 100.0\% with a window size of 4096. However, larger window sizes also increase inference time. With a chunk size of 512, increasing the window size from 512 to 4096 increases inference time from 528 to 646 seconds.

\textbf{Interaction between chunk size and window size}: The results suggest that chunk size and window size are interdependent and should be tuned together.
(1) For reliable performance, the chunk size should be smaller than the window size, ensuring that each chunk fits within the model's memory.
(2) For optimal efficiency, we fixed the chunk-to-window size ratio at 1:2. As shown in \Cref{fig:u}, inference time forms a U-shaped curve as chunk and window sizes vary. With small sizes, increasing the chunk size greatly improves efficiency with fewer iterations, but with large sizes, the cost of a larger window outweighs these gains. In our experiments, the best efficiency was achieved with a chunk size of 4096 and a window size of 8192.

\section{Conclusion}

Recurrent LLMs face considerable challenges in sustaining high performance on long-context tasks, limiting their applicability in real-world scenarios. In this paper, we propose a novel inference method, \emph{Smooth Reading}, which enhances the performance of Recurrent LLMs on long-context tasks. Our approach elevates their effectiveness to a level comparable to that of Self-Attention LLMs. This advancement positions Recurrent LLM as a more practical alternative to Self-Attention LLMs and expands the potential for future developments in Recurrent LLMs.

\bibliography{neurips_bib}
\bibliographystyle{abbrvnat}

\appendix
\newpage

\section{Technical Appendices and Supplementary Material}

\begin{figure}[t]
    \centering
    \includegraphics[width=.5\textwidth]{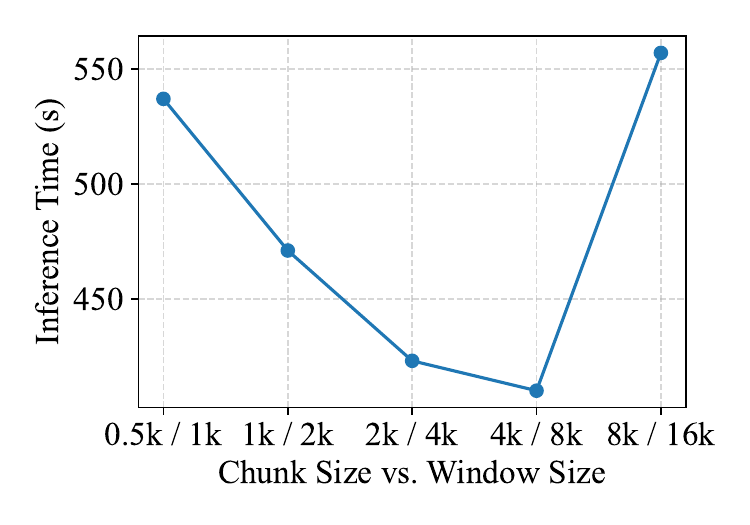}
    \caption{Inference Time Comparison with a Fixed Ratio of Chunk Size to Window Size on NIAH.}
    \label{fig:u}
\end{figure}

\subsection{Implementation Details}
\label{sec:implementation}

\begin{table}[b]
    \centering
    \caption{Configuration for Dataset Construction. ``Num'' indicates the number of samples, and ``Avg. Length'' indicates the average length of the samples. The raw datasets include HotpotQA \citep{HotpotQA}, NarrativeQA \citep{NarrativeQA}, GovReport \citep{gov_report}, QMSum \citep{qmsum}, TREC \citep{trec}, TriviaQA \citep{TriviaQA}, SAMSum \citep{samsum}, WikiSum \citep{wikisum}, and LCC \citep{LongCoder}.}
    \resizebox{\linewidth}{!}{
        \begin{tabular}{l|cccc|cc}
            \toprule
            Task                                & Raw Dataset & Teacher & Early Stop & Clean Metric    & Num    & Avg. Length \\
            \midrule
            \multirow{2}{*}{Question Answering} & HotpotQA    & LLM     & Yes        & F1              & 5,652  & 4,249       \\
                                                & NarrativeQA & LLM     & Yes        & F1              & 4,348  & 10,173      \\
            \midrule
            \multirow{2}{*}{Summarization}      & GovReport   & LLM     & No         & Rouge-L         & 8,114  & 12,411      \\
                                                & QMSum       & LLM     & No         & Rouge-L         & 742    & 15,245      \\
            \midrule
            \multirow{3}{*}{Few-shot}           & TREC        & LLM     & No         & Exact Match     & 3,259  & 7,759       \\
                                                & TriviaQA    & LLM     & No         & F1              & 3,370  & 13,942      \\
                                                & SAMSum      & LLM     & No         & Rouge-L         & 3,371  & 12,100      \\
            \midrule
            Passage Count                       & WikiSum     & Rule    & No         & Exact Match     & 3,333  & 15,777      \\
            \midrule
            Passage Retrieval                   & WikiSum     & LLM     & Yes        & Exact Match     & 3,333  & 8,254       \\
            \midrule
            Needle-in-Haystack-A                & HotpotQA    & Rule    & No         & Exact Match     & 1,667  & 20,117      \\
            \midrule
            Needle-in-Haystack-B                & HotpotQA    & Rule    & Yes        & Exact Match     & 1,667  & 10,403      \\
            \midrule
            Code Generation                     & LCC         & LLM     & No         & Edit Similarity & 10,000 & 3,331       \\
            \bottomrule
        \end{tabular}
    }
    \label{table:dataset-construction}
\end{table}

\subsubsection{Chunk Method}

\label{sec:chunk}

We employ a simple rule-based chunking method to divide the context into multiple chunks while maintaining semantic coherence. Our approach uses a hierarchical strategy based on a sequence of delimiters. For example, in a two-level chunking process, we first split the context into paragraphs using the delimiter ``\textbackslash n''. Adjacent paragraphs are then merged into chunks until they reach a specified maximum size. If a single paragraph exceeds the maximum size, it is further divided into sentences using the ``.'' delimiter, and the merging process is repeated at the sentence level.
In practice, we utilize a prioritized list of delimiters—``\textbackslash n\textbackslash n\textbackslash n'', ``\textbackslash n\textbackslash n'', ``\textbackslash n'', ``. '', ``.'', ``! '', ``? '', ``, '', ``; '', ``: '', `` -- '', and `` '' (space)—applied hierarchically to ensure appropriate chunk boundaries at multiple granularity levels. This method allows us to create context chunks that are both size-constrained and semantically coherent.

Since tokenization is slow and involves substantial overhead, we use a simple statistical method to estimate the number of tokens in a chunk: \(n_{token}\approx \mathrm{Int}(1.5\times n_{words})\). This approximation is based on the statistical observation that the average number of tokens is about 1.5 times the number of words.

\subsubsection{Dataset Construction}
\label{sec:dataset_construction_appendix}

\textbf{Teacher Model}: We employ either rule-based models or a SOTA LLM (deepseek-v3 \citep{deepseekv3}) as the teacher model, depending on task complexity. Rule-based models are used for simpler tasks (passage counting and NIAH), while the SOTA LLM handles more complex tasks (summarization and question answering). For LLM-based teaching, we use one-shot prompting with manually crafted examples to guide the generation of contextual summaries.
For generated contextual summaries, if reading continues, we append a ``<CONTINUE>'' token to the end of the generated text; otherwise, we append a ``<STOP>'' token.

\textbf{Maximum Chunk Size}: When using a SOTA LLM as the teacher model, we set a conservative chunk size of 512 tokens to improve accuracy. For rule-based models, the maximum chunk size varies between 128 and 4096 tokens to enhance generalization across different chunk sizes.

The configuration of our dataset construction is summarized in \Cref{table:dataset-construction}.

\subsubsection{Training Details}
\label{sec:training}
We use AdamW~\citep{adamw} with a learning rate of 4e-5 and a weight decay of 0.01. The batch size is set to 2, and the context length is set to 32k. We use a cosine learning rate decay schedule with warmup. As an exception, we use a learning rate of 1e-5 and a weight decay of 0.1 for models with 7B parameters, as we find this setting to be more effective.
We train our models with Xtuner~\citep{2023xtuner}. For models with 3B parameters, we train them for 1 epoch on one H800 GPU, with about 12 hours of training time.

\subsubsection{Evaluation Details}
\label{sec:eval}

We evaluate our models using the LongBench~\citep{longbench} and NIAH~\citep{ruler} benchmarks.

\textbf{LongBench}: LongBench comprises five main task categories, each with several sub-tasks:
\begin{itemize}[leftmargin=15pt]
    \item \textbf{Single-document question answering (SQA)}: NarrativeQA~\citep{NarrativeQA}, Qasper~\citep{Qasper}, MultiFieldQA~\citep{longbench}
    \item \textbf{Multi-document question answering (MQA)}: HotpotQA~\citep{HotpotQA}, 2WikiMultihopQA~\citep{2WikiMultihopQA}, MuSiQue~\citep{MuSiQue}
    \item \textbf{Summarization}: GovReport~\citep{gov_report}, QMSum~\citep{qmsum}, MultiNews~\citep{MultiNews}
    \item \textbf{Few-shot learning}: TREC~\citep{trec}, TriviaQA~\citep{TriviaQA}, SAMSum~\citep{samsum}
    \item \textbf{Synthetic tasks}: PassageCount~\citep{longbench}, PassageRetrieval~\citep{longbench}
    \item \textbf{Code generation}: LCC~\citep{LongCoder}, RepoBench-P~\citep{RepoBench}
\end{itemize}

\textbf{Needle-in-a-Haystack (NIAH)}: We use essays as haystacks, words as keys, and UUIDs as values.
We use only one needle in our experiments by default, but four needles in \Cref{tab:ruler}, as one needle is too simple to distinguish between models.

\textbf{Chunk Size during Evaluation}: During inference with \emph{Smooth Reading}, we set different chunk sizes according to the model type and task.
For Sliding-Window LLMs, we use a chunk size of 1024 for LongBench and 2048 for NIAH.
For RWKV-7, we use a chunk size of 512 for LongBench and 256 for NIAH.

\textbf{Inference Engine}: We use LMDeploy~\citep{2023lmdeploy}, which provides high performance for Self-Attention LLMs and supports Sliding-Window LLMs in interactive mode, maintaining hidden memory throughout inference. It strongly supports our \emph{Smooth Reading} inference.

\subsection{Comparison of the Length Extrapolation Ability of Sliding-Window LLM and RWKV-7}
\label{sec:rwkv-7}

\begin{figure}
    \centering
    \includegraphics[width=0.5\textwidth]{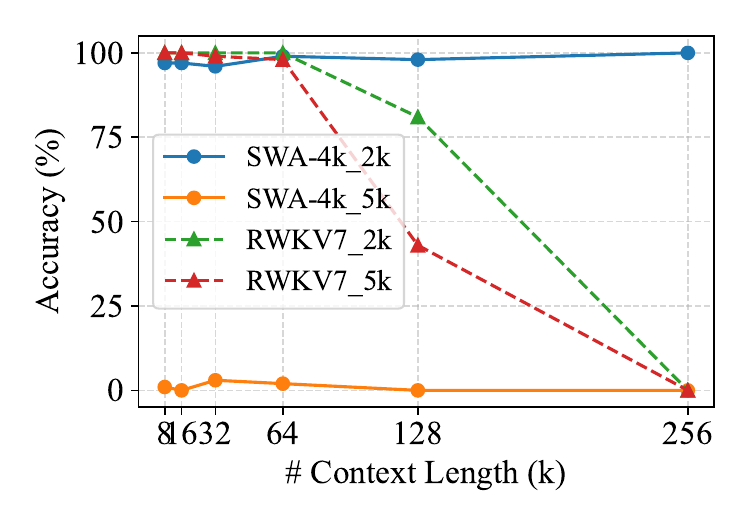}
    \caption{Comparison of Sliding-Window LLM and RWKV-7 on NIAH. The label for each line is "\{arch\}\_\{needle position\}". We compare two models: the Sliding-Window LLM with a 4k window size (denoted as SWA-4k) and RWKV-7. The needle position is set to the last 2048th and 5120th tokens. This figure shows that the Sliding-Window LLM has better length extrapolation ability, while RWKV-7 exhibits longer context memory within the training length but lower length extrapolation ability.}
    \label{fig:arch_compare}
    \vspace{-0.3cm}
\end{figure}

We evaluate the length extrapolation ability of RWKV-7 and Sliding-Window LLMs with One-Step inference using the NIAH benchmark, where the needle is placed at fixed positions. Specifically, we insert the needle at two locations: the last 2048th token (within the 4k token window of the Sliding-Window LLM) and the last 5120th token (outside this window). Both models are trained with a 32k context length using a One-Step style dataset, and the results are shown in \Cref{fig:arch_compare}.

\textbf{Sliding-Window LLMs}: We observe that Sliding-Window LLMs can recall the needle at the last 2048th token with nearly perfect accuracy, even when the context length is extended to 256k tokens. However, they fail to recall the needle at the last 5120th token, indicating an inability to retrieve information beyond their 4k window. Although performance is extremely low outside the window, these models demonstrate strong length extrapolation, as their accuracy within the window does not degrade when the context length exceeds the training length.

\textbf{RWKV-7}: In contrast, RWKV-7 performs well within the training context length, perfectly retrieving needles at both the last 2048th and 5120th tokens. However, it fails to extrapolate beyond the training context length: when the context is extended to 256k tokens, RWKV-7 cannot recall the needle even at the last 2048th token, indicating limited extrapolation ability. This observation is consistent with findings from~\citep{chenStuffedMambaState2024}.

\textbf{Further Discussion}: As shown in \Cref{tab:ruler}, with different inference methods, architectures exhibit varying performance rankings. Under One-Step inference, RWKV-7 outperforms Sliding-Window LLMs due to its superior long-range information extraction capability. However, when using \emph{Smooth Reading}, Sliding-Window LLMs surpass RWKV-7, benefiting from their strong length extrapolation ability. Consequently, different inference methods may require distinct optimization priorities in architectural design. We emphasize the importance of considering the inference method when designing model architectures.

In summary, Sliding-Window LLM exhibits better length extrapolation ability than RWKV-7, making them more suitable than RWKV-7 for our \emph{Smooth Reading} method.
As a comprehensive architectural comparison is beyond the scope of this work, we leave further analysis for future research.

\subsection{Additional Experiments on 7B Models}
\label{sec:7b}

\begin{table}[t]
    \centering

    \caption{Results on LongBench with Models with 7B Parameters. ``Infer'' indicates inference method, ``OS'' indicates One-Step inference, and ``SR'' indicates Smooth Reading.}
    \resizebox{\textwidth}{!}{
        \begin{tabular}{l|l|c|cccccc|c}
            \toprule
            Architecture                    & Infer & Model          & SQA            & MQA            & Summary        & FewShot        & Synthetic      & Code           & Avg            \\
            \midrule
            \multirow{2}{*}{Self-Attention} & OS    & Qwen-2.5-7B-OS & 34.90          & \textbf{56.72} & \textbf{32.07} & \textbf{72.77} & 57.75          & \textbf{73.41} & \textbf{54.60} \\
                                            & UR    & Qwen-2.5-7B-UR & 34.37          & 45.37          & 25.20          & 71.27          & \textbf{66.50} & 65.26          & 51.33          \\
            \midrule
            \multirow{2}{*}{Recurrent}      & OS    & SWA-7B-4k-OS   & 26.10          & 37.04          & 28.12          & 71.50          & 42.38          & 71.88          & 46.17          \\

                                            & SR    & SWA-7B-4k-SR   & \textbf{37.80} & 54.67          & 27.39          & 69.86          & 65.75          & 67.71          & 53.86          \\
            \bottomrule
        \end{tabular}
    }
    \label{tab:7b}
\end{table}

To further assess the scalability of our method, we conduct experiments using 7B-parameter models. The results, presented in \Cref{tab:7b}, show that our Recurrent LLM with \emph{Smooth Reading} achieves performance that is comparable to Self-Attention LLMs with One-Step inference and outperforms Self-Attention LLMs with Unsmooth Reading, as well as Sliding-Window LLMs with One-Step inference.
Specifically, the performance gap between SWA-7B-4k-SR and Qwen-2.5-7B-OS is less than 1\%, and SWA-7B-4k-SR outperforms Qwen-2.5-7B-UR and SWA-7B-4k-OS by 2.53\% and 7.69\%, respectively.
These findings demonstrate the effectiveness of our proposed method.

\subsection{Influence of Early Stopping on Performance}
\label{sec:early_stop}
\begin{table}[t]
    \centering
    \small
    \caption{Comparison of Performance with and without Early Stopping on NIAH.}
    \begin{tabular}{l|cccccc|c}
        \toprule
        Early Stop & 8k    & 16k    & 32k    & 64k   & 128k  & 256k   & Avg   \\
        \midrule
        without    & 99.80 & 100.00 & 100.00 & 99.80 & 99.60 & 100.00 & 99.87 \\
        with       & 99.60 & 99.60  & 99.80  & 99.80 & 99.60 & 99.80  & 99.70 \\
        \bottomrule
    \end{tabular}
    \label{tab:early_stop}
\end{table}

We further compare the performance of our method with and without early stopping by evaluating SWA-3B-4k-SR on NIAH, as presented in \Cref{tab:early_stop}.
The results show that early stopping has minimal impact on performance, with average accuracy exceeding 99\% in both scenarios.

\subsection{Comparison with Other Multiple-Step Inference Methods}
\label{sec:rag}
To assess the effectiveness of our approach, we compare \emph{Smooth Reading} with several other LLM inference methods. For a fair evaluation, we implement each method as follows:

\begin{itemize}[leftmargin=15pt]
    \item \textbf{RAG}: We utilize standard RAG models as introduced by~\citep{asai2023self}. For each query, Contriever-MS MARCO retrieves the top five documents from Wikipedia, using the official embeddings from the 2018 English Wikipedia. These retrieved passages form the long context. Note that this method incorporates external knowledge.
    \item \textbf{RAG+}: To avoid using external knowledge, we split the context into multiple passages for each question. Both the query and each passage are encoded using sentence embedding models~\citep{reimers2019sentence}. We compute the cosine similarity between the query and the passages, select the top three most relevant passages, and provide them as the context for answer generation.
    \item \textbf{CompACT}~\citep{CompAct}: We also compare with CompACT, a strong baseline. This method iteratively compresses the context based on the question, and the answer is generated from this compressed context. In our experiments, we adopt the off-the-shelf CompACT compressor.\footnote{\url{https://huggingface.co/cwyoon99/CompAct-7b}}
\end{itemize}

\begin{table}[t]
    \centering
    \caption{Comparison with More Multiple-Step Inference Methods on Question Answering Tasks.}
    \resizebox{\textwidth}{!}{
        \begin{tabular}{ll|l|cccc}
            \toprule
            Architecture                    & Infer   & Model                & HotpotQA       & MuSiQue        & 2WikiMQA       & TriviaQA       \\
            \midrule
            \multirow{3}{*}{Self-Attention} & RAG     & Qwen-2.5-3B-Instruct & 34.59          & 13.44          & 27.67          & 81.16          \\
                                            & RAG+    & Qwen-2.5-3B-Instruct & 36.91          & 19.92          & 35.47          & 82.15          \\
                                            & CompACT & Qwen-2.5-3B-Instruct & 36.57          & 18.35          & 39.95          & 79.69          \\
            \midrule
            Recurrent                       & SR      & SWA-4k               & \textbf{54.25} & \textbf{31.77} & \textbf{56.99} & \textbf{85.61} \\
            \bottomrule
        \end{tabular}
    }
    \label{tab:rag}
\end{table}

As shown in Table~\ref{tab:rag}, our \emph{Smooth Reading} approach consistently achieves the highest F1-scores across all evaluated datasets, demonstrating its superior capability in handling long-context passages. Compared to the compressor-based method CompACT, \emph{Smooth Reading} attains better results, which suggests that the compression process in CompACT may discard information important for accurate reasoning.

\subsection{Limitations}
\label{sec:limitations}

In this paper, we propose a novel inference method, \emph{Smooth Reading}, to enhance the performance of Recurrent LLMs on long-context tasks and curate a dataset specifically for training models on \emph{Smooth Reading}. As an early exploratory study, our main limitation is that the collected dataset is not generalizable across a wide range of tasks, as we focus only on two long-context benchmarks: NIAH and LongBench. Given that dataset collection is time-consuming and costly, we leave the development of a more comprehensive dataset for Recurrent LLMs to future work and encourage further research in this area.

\subsection{Broader Impact}
\label{sec:broader_impact}

\textbf{Ethical Concerns and Societal Impacts.} We declare that there are no conflicts of interest that could inappropriately influence our work. Our study does not involve human subjects, personal data collection, or experiments involving protected groups. All datasets used in this work are publicly available and widely adopted in the research community, such as LongBench and Needle-in-a-Haystack benchmarks. These datasets have been carefully curated to minimize biases and ethical risks.

To ensure fairness and transparency, we rigorously evaluated our methods across diverse data subsets, confirming that our findings are not affected by unintended bias. We also critically examine the limitations of our approach, avoiding overstatements of its capabilities. By leveraging open benchmarks, we enable independent validation and comparison with existing work.

\textbf{Positive Impacts.} Our work could benefit society by enabling more efficient processing of long documents (e.g., legal, medical) while reducing computational costs. We encourage follow-up studies to assess real-world impacts and to develop safeguards.

\end{document}